\documentclass[11pt,a4paper]{article}
\usepackage[hyperref]{emnlp2018}
\usepackage{times}
\usepackage{latexsym}
\usepackage{graphicx}
\usepackage{caption}
\usepackage{subcaption}
\usepackage{amsmath}
\usepackage{multirow}
\usepackage{booktabs}
\usepackage{forest}
\usepackage{float}
\forestset{
    nice empty nodes/.style={
        for tree={
            s sep=0.1em, 
            l sep=0.33em,
            inner ysep=0.4em, 
            inner xsep=0.05em,
            l=0,
            calign=midpoint,
            fit=tight,
            where n children=0{
               tier=word,
               minimum height=1.25em,
            }{},
            where n children=2{
               l-=1em,
            }{},
            parent anchor=south,
            child anchor=north,
            delay={if content={}{
                    inner sep=0pt,
                    edge path={\noexpand\path [\forestoption{edge}] 
                    			(!u.parent anchor) 
                               -- (.south)\forestoption{edge label};}
                }{}}
        },
    },
}

\usepackage{url}

\aclfinalcopy 

\usepackage[]{todonotes}  

\title{Grammar Induction with Neural Language Models:\\ An Unusual Replication}
% ...

\author{Phu Mon Htut$^1$\\
  AdeptMind Scholar \\
  {\tt pmh330@nyu.edu} \\\And
  Kyunghyun Cho$^{1,2}$ \\
  CIFAR Global Scholar \\
  {\tt kyunghyun.cho@nyu.edu} \\\And
  Samuel R. Bowman$^{1,2,3}$ \\
  {\tt bowman@nyu.edu} \\
  \AND
$^{1}$\normalfont Center for Data Science\\New York University\\60 Fifth Avenue\\New York, NY 10011\And
$^{2}$\normalfont Dept. of Computer Science\\New York University\\60 Fifth Avenue\\New York, NY 10011\And
$^{3}$\normalfont Dept. of Linguistics\\New York University\\10 Washington Place\\New York, NY 10003
  } 

\date{}

\begin{document}
\maketitle
\begin{abstract}
A substantial thread of recent work on \textit{latent tree learning} has attempted to develop neural network models with parse-valued latent variables and train them on non-parsing tasks, in the hope of having them discover interpretable tree structure. In a recent paper, \citet{shen2018neural} introduce such a model and report near-state-of-the-art results on the target task of language modeling, and the first strong latent tree learning result on  constituency parsing. 
In an attempt to reproduce these results, we discover issues that make the original results hard to trust, including tuning and even training on what is effectively the test set. Here, we attempt to reproduce these results in a fair experiment and to extend them to two new datasets. We find that the results of this work are robust: All variants of the model under study outperform all latent tree learning baselines, and perform competitively with symbolic grammar induction systems. We find that this model represents the first empirical success for latent tree learning, and that neural network language modeling warrants further study as a setting for grammar induction.

\end{abstract}

\section{Introduction and Background} 

Work on \textit{grammar induction} attempts to find methods for syntactic parsing that do not require expensive and difficult-to-design expert-labeled treebanks for training \citep{Charniak1992,Klein2001,Smith2005}. 
Recent work on \textit{latent tree learning} offers a new family of approaches to the problem \citep{YogatamaBDGL16,maillard2017jointly,choi2017unsupervised}. Latent tree learning models attempt to induce syntactic structure using the supervision from a downstream NLP task such as textual entailment.
Though these models tend to show good task performance, they are often not evaluated using standard parsing metrics, and \citet{Williams2018a} report that the parses they produce tend to be no better than random trees in a standard evaluation on the full Wall Street Journal section of the Penn Treebank \citep[WSJ;][]{DBLP:journals/coling/MarcusSM94}.

\begin{figure*}[!t]

	\centering
	\scalebox{0.7}{
	\begin{forest}
		shape=coordinate,
		where n children=0{
			tier=word
		}{},
		nice empty nodes
        [ [ [There] [ ['s] [ [ [nothing] [ [worth] [seeing] ] ] [ [in] [ [ [the] [tourist] ] [offices] ] ] ] ] ] [$.$] ]
	\end{forest}}
	\hspace{1.0em}
	\scalebox{0.7}{
	\begin{forest}
		shape=coordinate,
		where n children=0{
			tier=word
		}{},
		nice empty nodes
		[ [ [There] [ [ [ [ ['s] [nothing] ] [worth] ] [seeing] ] [ [in] [ [ [the] [tourist] ] [offices] ] ] ] ] [$.$] ]
\end{forest}}
	\vspace{0.3em}\\

	\vspace{1.0em}
	
	\scalebox{0.7}{
	\begin{forest}
		shape=coordinate,
		where n children=0{
			tier=word
		}{},
		nice empty nodes
        [ [ [ [The] [ [ [entire] [Minoan] ] [civilization] ] ] [ [was] [ [destroyed] [ [by] [ [ [a] [volcanic] ] [eruption] ] ] ] ] ] [$.$] ] 
	\end{forest}}\hspace{1.0em}
	\scalebox{0.7}{
	\begin{forest}
		shape=coordinate,
		where n children=0{
			tier=word
		}{},
		nice empty nodes
		[ [ [ [The] [ [entire] [Minoan] ] ] [civilization] ] [ [was] [ [ [destroyed] [ [by] [ [ [a] [volcanic] ] [eruption] ] ] ] [$.$] ] ] ] 
\end{forest}}
	\vspace{0.3em}\\
	
\caption{\label{fig:trees} \textit{Left} Parses from PRPN-LM trained on AllNLI. \textit{Right} Parses from PRPN-UP trained on AllNLI (stopping criterion: parsing).
We can observe that both sets of parses tend to have roughly reasonable high-level structure and tend to identify noun phrases correctly.
}
\end{figure*}
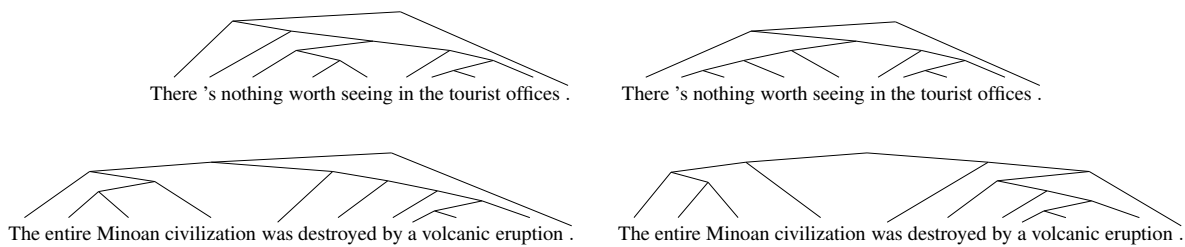

This paper addresses the Parsing-Reading-Predict Network \citep[PRPN;][]{shen2018neural}, which was recently published at ICLR, and which reports near-state-of-the-art results on language modeling and strong results on grammar induction, a first for latent tree models (though they do not use that term). PRPN is built around a substantially novel architecture, and uses convolutional networks with a form of structured attention \citep{KimDHR17} rather than recursive neural networks \citep{Goller1996,Socher2011} to evaluate and learn trees while performing straightforward backpropagation training on a language modeling objective. 
In this work, we aim to understand what the PRPN model learns that allows it to succeed, and to identify the conditions under which this success is possible.

Their experiments on language modeling and parsing are carried out using different configurations of the PRPN model, which were claimed to be optimized for the corresponding tasks. PRPN-LM is tuned for language modeling performance, and PRPN-UP for (unsupervised) parsing performance. In the parsing experiments, we also observe that the WSJ data is not split, such that the test data is used without parse information for training. This approach follows the previous works on grammar induction using non-neural models where the entire dataset is used for training \citep{Klein2001}. However, this implies that the parsing results of PRPN-UP may not be generalizable in the way usually expected of machine learning evaluation results. Additionally, it is not obvious that the model should be able to learn to parse reliably: (1) Since the parser is trained as part of a language model, it makes parsing decisions greedily and with \textit{no} access to any words to the right of the point where each parsing decision must be made \citep{DBLP:conf/acl/CollinsR04}; (2) As RNN language models are known to be insufficient for capturing syntax-sensitive dependencies \citep{DBLP:journals/tacl/LinzenDG16}, language modeling as the downstream task may not be well-suited to latent tree learning.

In this replication we train PRPN on two corpora: The full WSJ, a staple in work on grammar induction, and AllNLI, the concatenation of the Stanford Natural Language Inference Corpus \citep[SNLI; ][]{snli:emnlp2015} and the Multi-Genre NLI Corpus \citep[MultiNLI; ][]{WilliamsNB17}, which is used in other latent tree learning work for its non-syntactic classification labels for the task of textual entailment, and which we include for comparison. 
We then evaluate the constituency trees produced by these models on the WSJ test set, full WSJ10,\footnote{A standard processed subset of WSJ used in grammar induction in which the sentences contain no punctuation and no more than 10 words.} and the MultiNLI development set.

Our results indicate that PRPN-LM achieves better parsing performance than PRPN-UP on both WSJ and WSJ10 even though PRPN-UP was tuned---at least to some extent---for parsing. 
Surprisingly, a PRPN-LM model trained on the large out-of-domain AllNLI dataset achieves the best parsing performance on WSJ despite not being tuned for parsing. 
We also notice that vocabulary size affects the language modeling significantly---the perplexity gets higher as the vocabulary size increases.

Overall, 
despite the relatively uninformative experimental design used in \citet{shen2018neural}, we find that PRPN is an effective model. It outperforms all latent tree learning baselines by large margins on both WSJ and MultiNLI, and performs competitively with symbolic grammar induction systems on WSJ10, suggesting that PRPN in particular and language modeling in general are a viable setting for latent tree learning.

\section{Methods}

\begin{table*}[t]
\small
\setlength{\tabcolsep}{2.8 pt} 
\begin{center}
\begin{tabular}{llllccccccccc}
\toprule
 & \bf \multirow{3}{*}{\shortstack[l]{Training \\Data}} & \bf \multirow{3}{*}{\shortstack[l]{Stopping \\Criterion}} & \bf \multirow{3}{*}{\shortstack[l]{Vocab \\Size}} & \multicolumn{4}{c}{\bf Parsing F1} & \bf \multirow{3}{*}{\shortstack[l]{Depth \\WSJ}} & \multicolumn{4}{c}{\bf \multirow{2}{*}{\shortstack{Accuracy on WSJ by Tag}}} \\ 
\bf Model & & & & \multicolumn{2}{c}{\bf WSJ10} & \multicolumn{2}{c}{\bf WSJ } & & \bf \multirow{2}{*}{\shortstack{ADJP}} & \bf \multirow{2}{*}{\shortstack{NP}} & \bf \multirow{2}{*}{\shortstack{PP}} & \bf \multirow{2}{*}{\shortstack{INTJ}} \\
\bf  &   &  &  & \bf $\mu\:(\sigma)$ & \bf max & $\mu\:(\sigma)$ & \bf max  &  &   &   &   &   \\
 \midrule
PRPN-UP & AllNLI Train & UP & 76k & 67.5 (0.6) & 68.6 &   36.9 (0.6) & 38.0  & 5.8 &   29.3 & 62.0 & 31.6 & \it 0.0 \\
PRPN-UP &  AllNLI Train & LM & 76k & 66.3 (0.8) & 68.5 &  38.3 (0.5) & 39.8 & 5.8 & 28.7 & 65.5 & 32.7 &  \it 0.0 \\
PRPN-LM & AllNLI Train & LM & 76k & 52.4 (4.9) & 58.1 &  35.0 (5.4) &  \bf \underline{42.8} & 6.1 & \bf \underline{37.8} & 59.7 & \bf \underline{61.5} & \bf \underline{100.0} \\
\midrule
PRPN-UP & WSJ Full & UP & 15.8k & 64.7 (3.2)  & 70.9 &  26.4 (1.7) & 31.1 & 5.8 & 22.5  & 47.2 & 17.9 &  \it 0.0 \\
PRPN-UP  &  WSJ Full & LM & 15.8k & 64.3 (3.3) & 70.8  &  26.3 (1.8) & 30.8  & 5.8 & 22.7 & 46.6  & 17.8 &  \it 0.0 \\
PRPN-UP & WSJ Train & UP &  15.8k & 63.5 (3.5) & 70.7 &  26.2 (2.3) & 33.0  & 5.8 & 24.8 & 55.2 & 18.0 &  \it 0.0 \\
PRPN-UP  & WSJ Train & LM & 15.8k &  62.2 (3.9) & 70.3 &  26.0 (2.3) & 32.8 & 5.8 & 24.8 & 54.4 & 17.8 &  \it 0.0 \\
PRPN-LM & WSJ Train & LM & 10k & 70.5 (0.4) & \underline{71.3} &  37.4 (0.3) & 38.1 & 5.9 & 26.2 &  \bf \underline{63.9} & 24.4 &   \it 0.0 \\
PRPN-LM & WSJ Train & UP & 10k & 66.1 (0.5) & 67.2 &  33.4 (0.8) & 35.6 & 5.9 &  33.0 & 57.1 & 18.3 &  \it 0.0 \\
\midrule
300D ST-Gumbel & AllNLI Train  & NLI & -- & -- & -- & \it 19.0 (1.0) & \it 20.1 & -- & \it 15.6 & \it 18.8 & \it \it 9.9 & 59.4  \\
\hspace{1em} w/o Leaf GRU & AllNLI Train & NLI & -- & -- &  -- & 22.8 (1.6) & 25.0 & -- & 18.9 & 24.1 & \it 14.2 & 51.8  \\
300D RL-SPINN & AllNLI Train & NLI & -- &  -- & -- & \it 13.2  (0.0) & \it 13.2 & -- & \it 1.7 & \it 10.8 & \it 4.6 & 50.6  \\
\hspace{1em} w/o Leaf GRU & AllNLI Train & NLI & -- & -- & -- & \it 13.1 (0.1) & \it 13.2 & --  & \it 1.6 & \it 10.9 & \it 4.6 & 50.0 \\
\midrule 
CCM   &  WSJ10 Full & -- & -- & -- & 71.9 & -- & -- & -- & -- & -- & -- & -- \\
DMV+CCM  & WSJ10 Full & -- & -- & -- & 77.6 & -- &  -- & -- & -- & -- & -- & -- \\
UML-DOP &   WSJ10 Full & -- & -- & -- & \bf 82.9 & -- & -- & -- & -- & -- & -- & --  \\
\midrule
Random Trees & --  & -- & -- & -- & 34.7 & 21.3 (0.0) & 21.4 & 5.3 &17.4 & 22.3 & 16.0 & 40.4 \\
Balanced Trees & -- & -- & -- & -- & -- & 21.3 (0.0) & 21.3 & 4.6 & 22.1 & \textit{20.2} & \textit{9.3} & 55.9 \\
Left Branching & --  & -- & -- & \it 28.7 & 28.7 & 13.1 (0.0) & 13.1 & 12.4 & -- & -- & -- & --  \\
Right Branching &  -- & -- & -- & 61.7 & 61.7 & 16.5 (0.0) & 16.5  &  12.4 & -- & -- & -- & --  \\
\bottomrule 
\end{tabular}
\end{center}
\caption{\label{tab:wsj-table} Unlabeled parsing F1 results evaluated on full WSJ10 and WSJ test set broken down by training data and by early stopping criterion. 
The \textit{Accuracy} columns represent the fraction of ground truth constituents of a given type that correspond to constituents in the model parses. Italics mark results that are worse than the random baseline. Underlining marks the best results from our runs. Results with RL-SPINN and ST-Gumbel are from \citet{Williams2018a}, and are evaluated on the full WSJ. We run the model with 5 different random seeds to calculate the average F1. We use the model with the best F1 score to report ADJP, NP, PP, and INTJ.
WSJ10 baselines are from \citet[][CCM]{Klein2001}, \citet[][DMV+CCM]{Klein2005b}, and \citet[][UML-DOP]{Bod2006}. As the WSJ10 baselines are trained using additional information such as POS tags and dependency parser, they are not strictly comparable with the latent tree learning results.
}
\end{table*} 

PRPN consists of three components: (i) a  \textit{parsing network} that uses a two-layer convolution kernel to calculate the \textit{syntactic distance} between successive pairs of words, which can form an indirect representation of the constituency structure of the sentence, (ii) a recurrent \textit{reading network} that summarizes the current memory state based on all previous memory states and the implicit constituent structure, and (iii) a \textit{predict network} that uses the memory state to predict the next token. We refer readers to the appendix and the original work for details. 

We do not re-implement or re-tune PRPN, but rather attempt to replicate and understand the results of the work using the author's publicly available code.\footnote{\url{https://github.com/yikangshen/PRPN}} The experiments on language modeling and parsing are carried out using different configurations of the model, with substantially different hyperparameter values including the size of the word embeddings, the maximum sentence length, the vocabulary size, and the sizes of hidden layers. PRPN-LM is larger than PRPN-UP, with embedding layer that is 4 times larger and the number of units per layer that is 3 times larger. We use both versions of the model in all our experiments.

We use the 49k-sentence WSJ corpus in two settings. To replicate the original results, we re-run an experiment with no train/test split, and for a clearer picture of the model's performance, we run it again with the train (Section 0-21 of WSJ), validation (Section 22 of WSJ), and test (Section 23 of WSJ) splits. To compare PRPN to the models studied in \citet{Williams2018a}, we also retrain it on AllNLI. As the MultiNLI test set is not publicly available, we follow \citet{Williams2018a} and use the development set 
for testing. The parsing evaluation code in the original codebase does not support PRPN-LM, and we modify it in our experiments only to add this support.

For early stopping, we remove 10k random sentences from the MultiNLI training set and combine them with the SNLI development set to create a validation set. Our AllNLI training set contains 280.5K unique sentences (1.8M sentences in total including duplicate premise sentences),
 and covers six distinct genres of spoken and written English. We do not remove the duplicate sentences. We train the model for 100 epochs for WSJ and 15 epochs for AllNLI. We run the model five times with random initializations and average the results from the five runs.
The generated parses from the trained models with the best F1 scores and the pre-trained model that provides the highest F1 are available 
online.\footnote{\url{https://github.com/nyu-mll/PRPN-Analysis} 
}

\section{Experimental Results}

Table~\ref{tab:lm-table} shows our results for language modeling. PRPN-UP, configured as-is with parsing criterion and language modeling criterion, performs dramatically worse than the standard PRPN-LM (a vs.~d and e). However, this is not a fair comparison as the larger vocabulary gives PRPN-UP a harder task to solve.  Adjusting the vocabulary of PRPN-UP down to 10k to make a fairer comparison possible, the PPL of PRPN-UP improves significantly (c vs.~d), but not enough to match PRPN-LM (a vs.~c). We also observe that early stopping on parsing leads to incomplete training and a substantial decrease in perplexity (a vs.~b and d vs.~e). The models stop training at around the 13th epoch when we early-stop on parsing objective, while they stop training around the 65th epoch when we early-stop on language modeling objective. Both PRPN models trained on AllNLI do even worse (f and g), though the mismatch in vocabulary and domain may explain this effect. In addition, since it takes much longer to train PRPN on the larger AllNLI dataset, we train PRPN on AllNLI for only 15 epochs while we train the PRPN on WSJ for 100 epochs. Although the parsing objective converges within 15 epochs, we notice that language modeling perplexity is still improving. We expect that the perplexity of the PRPN models trained on AllNLI could be lower if we increase the number of training epochs.

\begin{table}[t]
\small
\centering
\setlength{\tabcolsep}{2 pt} 
\begin{center}
\begin{tabular}{clllrr}
\toprule
 & & \bf Training & \bf Stopping &  \bf Vocab & \bf PPL \\ 
& \bf Model & \bf Data & \bf Criterion & \bf Size & \bf Median \\
 \midrule
(a) & PRPN-LM & WSJ Train & LM & 10k & \bf 61.4  \\
(b) & PRPN-LM & WSJ Train & UP & 10k & 81.6 \\
(c) & PRPN-UP & WSJ Train & LM & 10k & 92.8 \\
(d) & PRPN-UP & WSJ Train & LM & 15.8k & 112.1 \\
(e) & PRPN-UP & WSJ Train & UP & 15.8k & 112.8 \\
\midrule
(f) & PRPN-UP & AllNLI Train & LM & 76k & 797.5 \\
(g) & PRPN-UP & AllNLI Train & UP & 76k & 848.9  \\
\bottomrule 
\end{tabular}
\end{center}
\caption{\label{tab:lm-table} Language modeling performance (perplexity) on the WSJ test set, broken down by training data used and by whether early stopping is done using the parsing objective (UP) or the language modeling objective (LM).  
}
\end{table}

\begin{table}
\small
\centering
\setlength{\tabcolsep}{2 pt} 
\begin{tabular}{lcccccccc}
\toprule
 & \bf Stopping &\multicolumn{3}{c}{\bf F1 wrt.} & \multicolumn{1}{c}{}\\
\bf Model & \bf Criterion & \multicolumn{1}{c}{\bf LB} & \multicolumn{1}{c}{\bf RB} & \multicolumn{1}{c}{\bf SP}  & \multicolumn{1}{c}{\bf Depth}\\
\midrule
300D SPINN & NLI & 19.3 & 36.9 & 70.2  & 6.2  \\
\hspace{1em} w/o Leaf GRU & NLI & 21.2 & 39.0  & 63.5  & 6.4  \\
300D SPINN-NC & NLI & 19.2 & 36.2 & 70.5  & 6.1  \\
\hspace{1em} w/o Leaf GRU & NLI & 20.6 & 38.9 & 64.1  & 6.3  \\
\midrule
300D ST-Gumbel & NLI & 32.6 & 37.5  & 23.7 & 4.1  \\
\hspace{1em} w/o Leaf GRU & NLI & 30.8  & 35.6 & 27.5 & 4.6  \\
300D RL-SPINN & NLI & 95.0  & 13.5  & 18.8  & 8.6  \\
\hspace{1em} w/o Leaf GRU & NLI & 99.1  & 10.7  & 18.1  & 8.6  \\
\midrule
PRPN-LM & LM & 25.6 & 26.9 & 45.7 & 4.9 \\
PRPN-UP & UP &  19.4   & 41.0 &  46.3 & 4.9 \\ 
PRPN-UP & LM & 19.9   & 37.4 & \bf 48.6  & 4.9\\  
\midrule
Random Trees & -- &  27.9  &  28.0  & 27.0  & 4.4  \\
Balanced Trees & -- & 21.7  &  36.8  & 21.3  & 3.9  \\
\bottomrule
\end{tabular} 
\caption{\label{tab:mnlitable} Unlabeled parsing F1 on the MultiNLI development set for models trained on AllNLI. \textit{F1 wrt.}~shows F1 with respect to strictly right- and left-branching (LB/RB) trees and with respect to the Stanford Parser (SP) trees supplied with the corpus; The evaluations of SPINN, RL-SPINN, and ST-Gumbel are from \citet{Williams2018a}. SPINN is a supervised parsing model, and the others are latent tree models. Median F1 of each model trained with 5 different random seeds is reported.
} 
\end{table}

Turning toward parsing performance, Table~\ref{tab:wsj-table} 
shows results with all the models under study, plus several baselines, on WSJ test set and full WSJ10. On full WSJ10, we reproduce the main parsing result of \citet{shen2018neural} with their UP model trained on WSJ without a data split. We also find the choice of parse quality as an early stopping criterion does not have a substantial effect and that training on the (unlabeled) test set does not give a significant improvement in performance. 
In addition and unexpectedly, we observe that PRPN-LM models achieve \textit{higher} parsing performance than PRPN-UP. This shows that any tuning done to separate PRPN-UP from PRPN-LM was not necessary, and more importantly, that the results described in the paper can be largely reproduced by a unified model in a fair setting. Moreover, the PRPN models trained on WSJ achieves comparable results with CCM \citep{Klein2001}. The PRPN models are outperformed by DMV+CCM\citep{Klein2005b}, and  UML-DOP\citep{Bod2006}. However, these models use additional information such as POS and dependency parser so they are not strictly comparable with the PRPN models.

Turning to the WSJ test set, the results look somewhat different: Although the differences in WSJ10 performance across models are small, the same is not true for the WSJ in terms of average F1. 
PRPN-LM outperforms all the other models on WSJ test set, even the potentially-overfit PRPN-UP model. Moreover, the PRPN models trained on the larger, out-of-domain AllNLI perform better than those trained on WSJ. Surprisingly, PRPN-LM tained  on out-of-domain AllNLI achieves the best F1 score on WSJ test set among all the models we experimented, even though its performance on WSJ10 is the lowest of all. This mean that PRPN-LM trained on AllNLI is strikingly good at parsing longer sentences though its performance on shorter sentences is worse than other models. 
Under all the configurations we tested, the PRPN model yields much better performance than the baselines from \citet[][called RL-SPINN]{YogatamaBDGL16} and \citet[][called ST-Gumbel]{choi2017unsupervised}, despite the fact that the model was tuned exclusively for WSJ10 parsing. This suggests that PRPN is consistently effective at latent tree learning. 

We also show detailed results for several specific constituent types, following \citet{Williams2018a}. We observe that the accuracy for NP (noun phrases) on the WSJ test set is above 46\% (Table \ref{tab:wsj-table}) for all PRPN models,  much higher than any of the baseline models. These runs also perform substantially better than the random baseline in the two other categories \citet{Williams2018a} report: ADJP (adjective phrases) and PP (prepositional phrases). However, as WSJ test set contains only one INTJ (interjection phrases), the results on INTJ are either 0.0\% or 100\%. 

In addition, Table~\ref{tab:mnlitable} shows that the PRPN-UP models achieve the median parsing F1 scores of 46.3 and 48.6 respectively on the MultiNLI dev set while PRPN-LM performs the median F1 of 45.7; setting the state of the art in parsing performance on this dataset among latent tree models by a large margin. We conclude that PRPN does acquire some substantial knowledge of syntax, and that this knowledge agrees with Penn Treebank (PTB) grammar significantly better than chance. 

Qualitatively, the parses produced by most of the best performing PRPN models
are relatively balanced (F1 score of 36.5 w.r.t balanced trees) and tend toward right branching (F1 score of 42.0 with respect to balanced trees). They are also shallower than average ground truth PTB parsed trees. These models can parse short sentences relatively well, as shown by their high WSJ10 performance.

For a large proportion of long sentences, 
most of the best performing models can produce reasonable constituents (Table ~\ref{tab:wsj-table}). The best performing model, PRPN-LM trained on AllNLI, achieves the best accuracy at identifying ADJP (adjective phrases), PP (prepositional phrases), and INTJ (interjection phrases) constituents, and a high accuracy on NP (noun phrases). In a more informal inspection, we also observe that our best PRPN-LM and PRPN-UP runs are fairly good at pairing determiners with NPs as we can observe in Figure~\ref{fig:trees}). 
Although lower level tree constituents appear random in many cases for both PRPN-LM and PRPN-UP, the intermediate and higher-level constituents are generally reasonable. For example, in Figure~\ref{fig:trees}, although the parse for lower level constituents like \textit{The entire Minoan} seem random, the higher-level constituents, such as \textit{The entire Minoan civilization} and \textit{nothing worth seeing in the tourist offices}, are reasonable.

\section{Conclusion}
\label{sec:conclusion}
In our attempt to replicate the grammar induction results reported in \citet{shen2018neural}, we find several experimental design problems that make the results difficult to interpret. However, in experiments and analyses going well beyond the scope of the original paper, we find that the PRPN model presented in that work is nonetheless robust. It represents a viable method for grammar induction and the first clear success for latent tree learning with neural networks, and we expect that it heralds further work on language modeling as a tool for grammar induction research. 

\section*{Acknowledgments}
This project has benefited from financial support to SB by Google and Tencent Holdings, and was partly supported by Samsung Electronics (Improving Deep Learning using Latent Structure). We thank Adina Williams, Katharina Kann, Ryan Cotterell, and the anonymous reviewers for their helpful comments and suggestions, and NVIDIA for their support.

\bibliography{emnlp2018}

\begin{thebibliography}{19}
\expandafter\ifx\csname natexlab\endcsname\relax\def\natexlab#1{#1}\fi

\bibitem[{Bod(2006)}]{Bod2006}
Rens Bod. 2006.
\newblock {An All-Subtrees Approach to Unsupervised Parsing}.
\newblock \emph{Proceedings of the 21st International Conference on
  Computational Linguistics and the 44th annual meeting of the Association for
  Computational Linguistics}, pages 865--872.

\bibitem[{Bowman et~al.(2015)Bowman, Angeli, Potts, and
  Manning}]{snli:emnlp2015}
Samuel~R. Bowman, Gabor Angeli, Christopher Potts, and Christopher~D. Manning.
  2015.
\newblock A large annotated corpus for learning natural language inference.
\newblock In \emph{Proceedings of the Conference on Empirical Methods in
  Natural Language Processing (EMNLP)}. Association for Computational
  Linguistics.

\bibitem[{Charniak and Carroll(1992)}]{Charniak1992}
Eugene Charniak and Glen Carroll. 1992.
\newblock {Two experiments on learning probabilistic dependency grammars from
  corpora}.
\newblock In \emph{Proceedings of the AAAI Workshop on Statistically-Based NLP
  Techniques}, page 1–13.

\bibitem[{Choi et~al.(2018)Choi, Yoo, and Lee}]{choi2017unsupervised}
Jihun Choi, Kang~Min Yoo, and Sang-goo Lee. 2018.
\newblock Learning to compose task-specific tree structures.
\newblock In \emph{Proceedings of the Thirty-Second Association for the
  Advancement of Artificial Intelligence Conference on Artificial Intelligence
  (AAAI-18)}, volume~2.

\bibitem[{Collins and Roark(2004)}]{DBLP:conf/acl/CollinsR04}
Michael Collins and Brian Roark. 2004.
\newblock Incremental parsing with the perceptron algorithm.
\newblock In \emph{Proceedings of the 42nd Annual Meeting of the Association
  for Computational Linguistics, 21-26 July, 2004, Barcelona, Spain.}, pages
  111--118.

\bibitem[{Goller and Kuchler(1996)}]{Goller1996}
Christoph Goller and Andreas Kuchler. 1996.
\newblock Learning task-dependent distributed representations by
  backpropagation through structure.
\newblock In \emph{Proceedings of International Conference on Neural Networks
  (ICNN'96)}.

\bibitem[{Hochreiter and Schmidhuber(1996)}]{Hochreiter1996}
Sepp Hochreiter and J{\"{u}}rgen Schmidhuber. 1996.
\newblock {Long Short Term Memory}.
\newblock \emph{Memory}, (1993):1--28.

\bibitem[{Kim et~al.(2017)Kim, Denton, Hoang, and Rush}]{KimDHR17}
Yoon Kim, Carl Denton, Luong Hoang, and Alexander~M. Rush. 2017.
\newblock Structured attention networks.

\bibitem[{Klein and Manning(2002)}]{Klein2001}
Dan Klein and Christopher~D. Manning. 2002.
\newblock {A generative constituent-context model for improved grammar
  induction}.
\newblock In \emph{Proceedings of the 40th Annual Meeting on Association for
  Computational Linguistics - ACL '02}, page 128.

\bibitem[{Klein and Manning(2005)}]{Klein2005b}
Dan Klein and Christopher~D. Manning. 2005.
\newblock {Natural language grammar induction with a generative
  constituent-context model}.
\newblock \emph{Pattern Recognition}, 38(9):1407--1419.

\bibitem[{Linzen et~al.(2016)Linzen, Dupoux, and
  Goldberg}]{DBLP:journals/tacl/LinzenDG16}
Tal Linzen, Emmanuel Dupoux, and Yoav Goldberg. 2016.
\newblock Assessing the ability of lstms to learn syntax-sensitive
  dependencies.
\newblock \emph{{TACL}}, 4:521--535.

\bibitem[{Maillard et~al.(2017)Maillard, Clark, and
  Yogatama}]{maillard2017jointly}
Jean Maillard, Stephen Clark, and Dani Yogatama. 2017.
\newblock Jointly learning sentence embeddings and syntax with unsupervised
  {T}ree-{LSTM}s.
\newblock {a}rXiv preprint 1705.09189.

\bibitem[{Marcus et~al.(1993)Marcus, Santorini, and
  Marcinkiewicz}]{DBLP:journals/coling/MarcusSM94}
Mitchell~P. Marcus, Beatrice Santorini, and Mary~Ann Marcinkiewicz. 1993.
\newblock Building a large annotated corpus of english: The penn treebank.
\newblock \emph{Computational Linguistics}, 19(2):313--330.

\bibitem[{Shen et~al.(2018)Shen, Lin, wei Huang, and
  Courville}]{shen2018neural}
Yikang Shen, Zhouhan Lin, Chin wei Huang, and Aaron Courville. 2018.
\newblock Neural language modeling by jointly learning syntax and lexicon.
\newblock In \emph{International Conference on Learning Representations}.

\bibitem[{Smith and Eisner(2005)}]{Smith2005}
Noah~A. Smith and Jason Eisner. 2005.
\newblock {Guiding unsupervised grammar induction using contrastive
  estimation}.
\newblock In \emph{Proceedings of IJCAI Workshop on Grammatical Inference
  Applications}, pages 73--82.

\bibitem[{Socher et~al.(2011)Socher, Lin, Ng, and Manning}]{Socher2011}
Richard Socher, Cliff Chiung-Yu Lin, Andrew Ng, and Chris Manning. 2011.
\newblock {Parsing Natural Scenes and Natural Language with Recursive Neural
  Networks}.
\newblock In \emph{Proceedings of the 28th International Conference on Machine
  Learning}, pages 129--136.

\bibitem[{Williams et~al.(2018{\natexlab{a}})Williams, Drozdov, and
  Bowman}]{Williams2018a}
Adina Williams, Andrew Drozdov, and Samuel~R. Bowman. 2018{\natexlab{a}}.
\newblock Do latent tree learning models identify meaningful structure in
  sentences?
\newblock \emph{Transactions of the Association for Computational Linguistics
  (TACL)}.

\bibitem[{Williams et~al.(2018{\natexlab{b}})Williams, Nangia, and
  Bowman}]{WilliamsNB17}
Adina Williams, Nikita Nangia, and Samuel~R. Bowman. 2018{\natexlab{b}}.
\newblock A broad-coverage challenge corpus for sentence understanding through
  inference.
\newblock In \emph{Proceedings of the North American Chapter of the Association
  for Computational Linguistics (NAACL)}.

\bibitem[{Yogatama et~al.(2017)Yogatama, Blunsom, Dyer, Grefenstette, and
  Ling}]{YogatamaBDGL16}
Dani Yogatama, Phil Blunsom, Chris Dyer, Edward Grefenstette, and Wang Ling.
  2017.
\newblock {Learning to Compose Words into Setences with Reinforcement
  Learning}.
\newblock \emph{Proceedings of the International Conference on Learning
  Representations}, pages 1--17.

\end{thebibliography}
\bibliographystyle{acl_natbib_nourl}


\newpage
\appendix
\setcounter{figure}{0} \renewcommand{\thefigure}{A.\arabic{figure}}

\newpage
\section{Parsing-Reading-Predict Network}
\label{sec:appendix}
The forward pass of PRPN is described here. Parsing-Reading-Predict Network contains three components.
\subsection{Parsing Network}
The syntactic distance between a given token represented as word embedding $e_{i}$ and the previous token $e_{i-1}$ is calculated by convolution kernel over a set of previous tokens $e_{i-L}$, $e_{i-L+1}$,..., $e_{i}$. Mathematically, syntactic distance $d_{i}$ between $e_{i-1}$ and $e_{i}$ is computed as:

\begin{equation} \label{conv_kernel_1}
h_i = \mathrm{ReLU}(W_c \left[ \begin{matrix} e_{i-L} \\ e_{i-L+1} \\ ... \\ e_i \end{matrix} \right] + b_c)
\end{equation}
\begin{equation} \label{conv_kernel_2}
d_i = \mathrm{ReLU} \left(W_d h_i + b_d\right)
\end{equation}
where $W_c$, $b_c$ are the kernel parameters. $W_d$ and $b_d$ can be seen as another convolutional kernel with window size 1, convolved over $h_i$'s. The kernel window size $L$, that indicates how far back into the history node $e_i$ can reach while computing its syntactic distance $d_i$, is called the \emph{look-back range}. For the tokens in the beginning of the sequence, $L-1$ zero vectors are padded to the front of the sequence. This will produce $K-1$ distances for sequence length of $K$. 

To determine the closest word $x_j$ that has larger syntactic relationship than $d_j$ for time step $t$, $\alpha_{j}^{t}$ is defined as:
\begin{equation}  \label{soft_alpha}
\alpha_j^t = \frac{\mathrm{hardtanh} \left( (d_t - d_{j}) \cdot \tau \right) + 1}{2} 
\end{equation}
where $\tau$ is the temperature parameter that controls the sensitivity of $\alpha_j^t$ to the differences between distances.

The soft gate values that will be used for language modeling are then computed as:
\begin{equation} \label{gate_equal_multialpha}
g_{i}^{t} = \mathbf{P}(l_{t}\leq i) = \prod_{j=i+1}^{t-1}\alpha_j^{t}
\end{equation}

\subsection{Reading Network}
The reading network uses Long Short-Term Memory-Network that maintains two sets of vectors as the memory states: a hidden tape $H_{t-1} = \left( h_{t-N_m}, ..., h_{t-1} \right)$, and a memory tape $C_{t-1} = \left( c_{t-L}, ..., c_{t-1} \right)$, where $N_m$ is the upper bound for the memory span. Hidden state $m_i$ is  represented by a tuple of two vectors $(h_i, c_i)$. 

At each step, the reading network links the current token to all the previous tokens that are syntactically similar:
\begin{align}
k_t &= W_h h_{t-1} + W_x x_t \\
\tilde{s}_i^t &= \mathrm{softmax}(\frac{h_i k_t^{\mathrm{T}}}{\sqrt{\delta_k}})
\end{align}
where, $\delta_k$ is the dimension of the hidden state. 
The structured intra-attention weight is defined based on the gates in Eq.\ref{gate_equal_multialpha}:
\begin{align}
s_i^t &= \frac{g_i^t \tilde{s}_i^t}{\sum_i g_i^t}
\end{align}
An adaptive summary vector for the previous hidden tape and memory denoted by $\tilde { h } _{ t }$ and $\tilde { c } _{ t }$ are computed as:
\begin{equation}
\left[ \begin{matrix} \tilde { h } _{ t } \\ \tilde { c } _{ t } \end{matrix} \right] = \sum _{ i=1 }^{ t-1 } s_{ i }^{ t }\cdot m_i = \sum _{ i=1 }^{ t-1 } s_{ i }^{ t }\cdot \left[ \begin{matrix} h_i \\ c_i \end{matrix} \right] 
\end{equation}

The Reading Network then takes $x_t$, $\tilde { c } _{ t }$ and $\tilde { h } _{ t }$ as input, computes the values of $c_t$ and $h_t$ by the LSTM recurrent update \citep{Hochreiter1996}.
Then the \textit{write} operation concatenates $h_t$ and $c_t$ to the end of hidden and memory tape.

\subsection{Predict Network}
Predict Network models the probability distribution of next word $x_{t+1}$, based on hidden states $m_0,...,m_t$, and gates $g_0^{t+1},...,g_{t}^{t+1}$:

\begin{align}
p(x_{t+1}|x_{t<t+1}) &\approx p(x_{t+1};f(m_{t<t+1},g_{t<t+1}^{t+1})) \label{eq_cond}
\end{align}

Since the model cannot observe $x_{t+1}$ at time step $t$, a temporary estimation of $d_{t+1}$ is computed using $x_{t-L},...,x_t$:
\begin{equation} \label{eq_g_estimate}
	d'_{t+1}=\mathrm{ReLU}(W'_d h_t + b'_d)
\end{equation}
The corresponding $\{\alpha^{t+1}\}$ and $\{g_i^{t+1}\}$ for Eq.\ref{eq_cond} are computed after that.
$f(\cdot)$ function is parameterized as:
\begin{align}
f(m_{t},...,m_t,g_0^{t+1},...,g_{t}^{t+1}) &= \hat{f}([h_{l:t-1},h_{t}]) \label{eq_pred}
\end{align}
where $h_{l:t-1}$ is an adaptive summary of $h_{l_{t+1} \leq i \leq t-1}$, output by structured attention controlled by $g_0^{t+1},...,g_{t-1}^{t+1}$. $\hat{f}(\cdot)$ could be a simple feed-forward MLP, or more complex architecture, like ResNet, to add more depth to the model.

\subsection{Sample Parses from the model with the best F1 score}
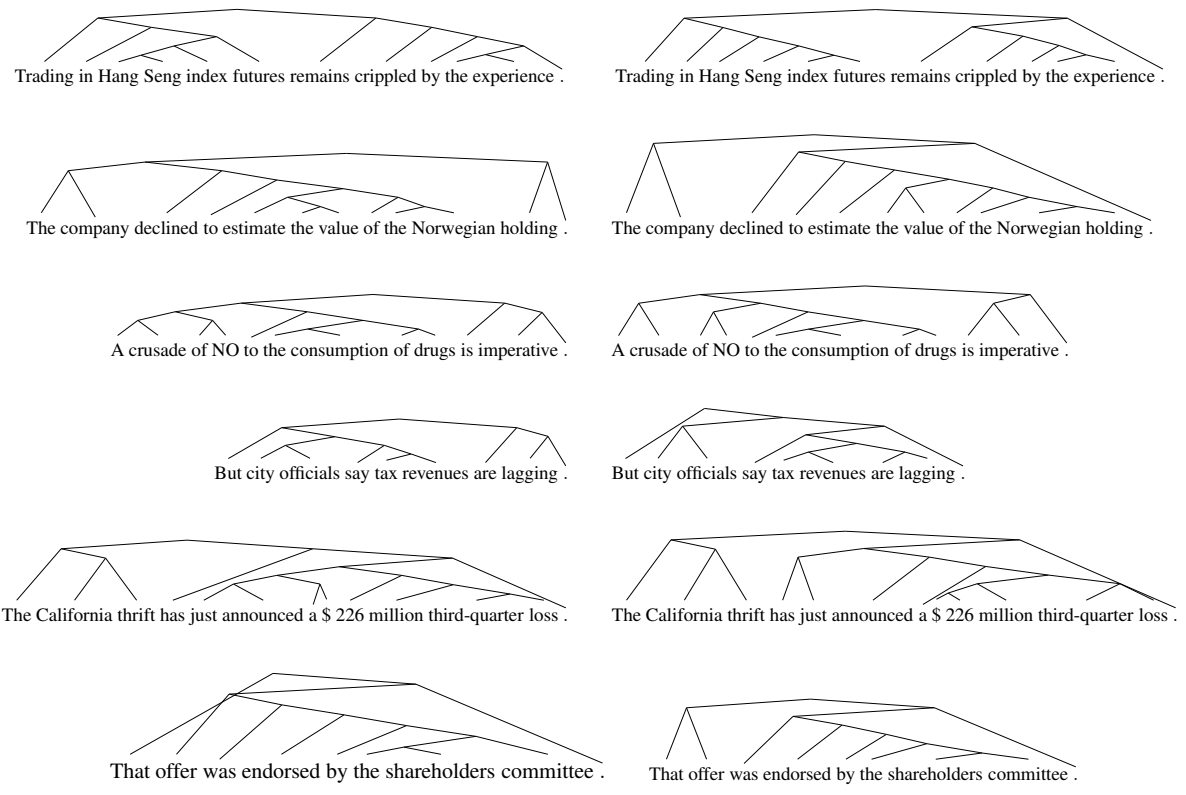
\begin{figure*}[!t]

	\centering
	\scalebox{0.65}{
	\begin{forest}
		shape=coordinate,
		where n children=0{
			tier=word
		}{},
		nice empty nodes
        [ [ [Trading] [ [in] [ [ [ [Hang] [Seng] ] [index] ] [futures] ] ] ] [ [remains] [ [crippled] [ [by] [ [ [the] [experience] ] [$.$] ] ] ] ] ]
	\end{forest}}
	\hspace{1.0em}
	\scalebox{0.65}{
	\begin{forest}
		shape=coordinate,
		where n children=0{
			tier=word
		}{},
		nice empty nodes
		[ [ [Trading] [ [in] [ [Hang] [ [Seng] [ [index] [futures] ] ] ] ] ] [ [ [remains] [ [crippled] [ [by] [ [the] [experience] ] ] ] ] [$.$] ] ]
\end{forest}}
	\vspace{0.3em}\\

	\vspace{1.0em}
	
	\scalebox{0.65}{
	\begin{forest}
		shape=coordinate,
		where n children=0{
			tier=word
		}{},
		nice empty nodes
        [ [ [ [The] [company] ] [ [declined] [ [to] [ [ [estimate] [ [the] [value] ] ] [ [of] [ [the] [Norwegian] ] ] ] ] ] ] [ [holding] [$.$] ] ]
	\end{forest}}\hspace{1.0em}
	\scalebox{0.65}{
	\begin{forest}
		shape=coordinate,
		where n children=0{
			tier=word
		}{},
		nice empty nodes
		[ [ [The] [company] ] [ [ [declined] [ [to] [ [estimate] [ [ [the] [value] ] [ [of] [ [the] [ [Norwegian] [holding] ] ] ] ] ] ] ] [$.$] ] ] 
\end{forest}}
	\vspace{0.3em}\\
    
    \vspace{1.0em}
	
	\scalebox{0.65}{
	\begin{forest}
		shape=coordinate,
		where n children=0{
			tier=word
		}{},
		nice empty nodes
        [ [ [ [ [A] [crusade] ] [ [of] [NO] ] ] [ [to] [ [ [the] [consumption] ] [ [of] [drugs] ] ] ] ] [ [is] [ [imperative] [$.$] ] ] ]
	\end{forest}}\hspace{1.0em}
	\scalebox{0.65}{
	\begin{forest}
		shape=coordinate,
		where n children=0{
			tier=word
		}{},
		nice empty nodes
		[ [ [ [A] [crusade] ] [ [ [of] [NO] ] [ [to] [ [ [the] [consumption] ] [ [of] [drugs] ] ] ] ] ] [ [ [is] [imperative] ] [$.$] ] ]
\end{forest}}
	\vspace{0.3em}\\
	
    \vspace{1.0em}
	
	\scalebox{0.65}{
	\begin{forest}
		shape=coordinate,
		where n children=0{
			tier=word
		}{},
		nice empty nodes
        [ [ [But] [ [ [city] [officials] ] [ [say] [ [tax] [revenues] ] ] ] ] [ [are] [ [lagging] [$.$] ] ] ] 
	\end{forest}}\hspace{1.0em}
	\scalebox{0.65}{
	\begin{forest}
		shape=coordinate,
		where n children=0{
			tier=word
		}{},
		nice empty nodes
		[ [But] [ [ [city] [officials] ] [ [ [say] [ [ [tax] [revenues] ] [ [are] [lagging] ] ] ] [$.$] ] ] ]
\end{forest}}
\vspace{0.3em}\\
	
    \vspace{1.0em}
	
	\scalebox{0.65}{
	\begin{forest}
		shape=coordinate,
		where n children=0{
			tier=word
		}{},
		nice empty nodes
        [ [ [The] [ [California] [thrift] ] ] [ [has] [ [ [ [ [just] [announced] ] [ [a] [\$] ] ] [ [226] [ [million] [ [third-quarter] [loss] ] ] ] ] [$.$] ] ] ]
	\end{forest}}\hspace{1.0em}
	\scalebox{0.65}{
	\begin{forest}
		shape=coordinate,
		where n children=0{
			tier=word
		}{},
		nice empty nodes
		[ [ [The] [ [California] [thrift] ] ] [ [ [ [has] [just] ] [ [announced] [ [a] [ [ [ [\$] [226] ] [million] ] [ [third-quarter] [loss] ] ] ] ] ] [$.$] ] ]
\end{forest}}
\vspace{0.3em}\\
	
    \vspace{1.0em}
	
	\scalebox{0.75}{
	\begin{forest}
		shape=coordinate,
		where n children=0{
			tier=word
		}{},
		nice empty nodes
        [ [That] [ [ [offer] [ [was] [ [endorsed] [ [by] [ [ [the] [shareholders] ] [committee] ] ] ] ] ] [$.$] ] ]
	\end{forest}}\hspace{1.0em}
	\scalebox{0.65}{
	\begin{forest}
		shape=coordinate,
		where n children=0{
			tier=word
		}{},
		nice empty nodes
		[ [ [That] [offer] ] [ [ [was] [ [endorsed] [ [by] [ [the] [ [shareholders] [committee] ] ] ] ] ] [$.$] ] ]
\end{forest}}

\caption{\label{fig:sample-trees} \textit{Left} Parses from PRPN-LM trained on AllNLI (stopping criterion: language modeling). \textit{Right} Ground truth parses from Penn Treebank.
}
\end{figure*}

In Figure~\ref{fig:sample-trees}, we report a few example parses from the model with the best F1 score (PRPN-UP trained on AllNLI) among our experiments, compared with the ground truth PTB parses.

 \end{document}